\def\year{2022}\relax
\documentclass[letterpaper]{article} % DO NOT CHANGE THIS
\usepackage{aaai22}  % DO NOT CHANGE THIS
\usepackage{times}  % DO NOT CHANGE THIS
\usepackage{helvet}  % DO NOT CHANGE THIS
\usepackage{courier}  % DO NOT CHANGE THIS
\usepackage[hyphens]{url}  % DO NOT CHANGE THIS
\usepackage{graphicx} % DO NOT CHANGE THIS
\usepackage{natbib}  % DO NOT CHANGE THIS AND DO NOT ADD ANY OPTIONS TO IT
\usepackage{caption} % DO NOT CHANGE THIS AND DO NOT ADD ANY OPTIONS TO IT
\usepackage{multirow}
\DeclareCaptionStyle{ruled}{labelfont=normalfont,labelsep=colon,strut=off} % DO NOT CHANGE THIS
\usepackage{algorithm}
\usepackage{algorithmic}

\usepackage{newfloat}
\usepackage{listings}
\floatstyle{ruled}
\newfloat{listing}{tb}{lst}{}
\floatname{listing}{Listing}
\title{The (In)Effectiveness of Intermediate Task Training \\ For Domain Adaptation and Cross-Lingual Transfer Learning}
\title{The (In)Effectiveness of Intermediate Task Training \\ For Domain Adaptation and Cross-Lingual Transfer Learning}
\author {
    Sovesh Mohapatra\textsuperscript{\rm 1,$\dagger$}, 
    Somesh Mohapatra \textsuperscript{\rm 2,$\dagger$}
}
\affiliations {
    \textsuperscript{\rm 1} University of Massachusetts Amherst\\
    \textsuperscript{\rm 2} Massachusetts Institute of Technology\\
    soveshmohapa@umass.edu, someshm@mit.edu \\
    \textsuperscript{$\dagger$}Both authors contributed equally to the work.
}

\usepackage{bibentry}
\begin{document}

\maketitle

\begin{abstract}
Transfer learning from large language models (LLMs) has emerged as a powerful technique to enable knowledge-based fine-tuning for a number of tasks, adaptation of models for different domains and even languages. However, it remains an open question, if and when transfer learning will work, i.e. leading to positive or negative transfer. In this paper, we analyze the knowledge transfer across three natural language processing (NLP) tasks - text classification, sentimental analysis, and sentence similarity, using three LLMs - BERT, RoBERTa, and XLNet -  and analyzing their performance, by fine-tuning on target datasets for domain and cross-lingual adaptation tasks, with and without an intermediate task training on a larger dataset. Our experiments showed that fine-tuning without an intermediate task training can lead to a better performance for most tasks, while more generalized tasks might necessitate a preceding intermediate task training step. We hope that this work will act as a guide on transfer learning to NLP practitioners.
\end{abstract}

\section{Introduction}
\noindent Knowledge-based transfer learning leverages zero or few-shot learning from a pre-trained model to predict for a range of similar tasks \cite{you2020co, raffel2020exploring, houlsby2019parameter}. The ability to use a pre-trained model, as-is or with very limited training, has proposed a very lucrative opportunity, as compared to training from scratch for every single task \cite{pan2020transfer, day2017survey} The applications of transfer learning have ranged from NLP to image, and even video tasks \cite{kim2020effectiveness, salza2022effectiveness, bengio2012deep}. 

In recent works, people have applied transfer learning to a range of NLP tasks, observing mixed results, both positive and negative transfer \cite{zhang2022survey}. \citet{pruksachatkun2020intermediate} showed how transfer learning with intermediate task training could affect a number of target and probing English-language NLP tasks. In most cases, positive transfer from LLMs, such as BERT, has been noted for similar language NLP tasks, like hate speech classification \cite{mozafari2019bert}, propaganda detection \cite{vlad2019sentence}, and biomedical NLP tasks \cite{peng2019transfer}. Negative transfer has been shown in attempts to transfer an English Part-of-Speech (POS) tagger to a Hindi corpus \cite{dell2009ensemble, rayson2007tagging}, and other NLP tasks \cite{wang2019transferable}. 

Transfer learning for domain adaptation has been widely studied and applied across language and medical fields \cite{xu2020transfer, ghafoorian2017transfer, kouw2018introduction}. \citet{savini2022intermediate} showed how intermediate task training on sarcasm helped in transfer learning, similar to \citet{felbo2017using} and \citet{buaroiu2022automatic}. However, in another domain adaptation task, \citet{meftah2021hidden} showed that knowledge transfer between related seemingly similar domains like news and tweets resulted in negative transfer, probing the results using both quantitative and qualitative methods.

Cross-lingual tasks are another area where transfer learning strategies have shown a lot of potential \cite{ahmad2020borrow, luo2021cross}. \citet{chen2018multi} have shown that when language-invariant and language-specific features are coupled at the instance level.

In this work, we analyze the effect of intermediate task training on a larger dataset for three different NLP tasks - text classification, sentiment analysis, and sentence similarity - and evaluate three language models - BERT, RoBERTa, and XLNet. For each NLP task, we have one domain adaptation and another cross-lingual task. In total, we have eighteen experiments on a range of NLP tasks.  

\section{Methodology}
Here, we present an overview of our methodology, including information on transfer learning for intermediate task training, and domain adaptation, and cross-lingual fine-tuning and evaluation for the NLP tasks, and the datasets.

In each of the following tasks, both intermediate task training and fine-tuning were performed by training over 70\% of the dataset, and evaluated on the remaining 30\%. For the intermediate task training, each pre-trained LLM was trained for 100 epochs using the large dataset. For fine-tuning after and without intermediate task training, transfer learning to the target dataset was performed by training for 10 epochs. In both cases of transfer learning, all the model weights were updated, or none of the layers were frozen.

Model instances for LLMs, BERT, RoBERTa, and XLNet, were obtained from the respective GitHub repositories \cite{qiu2020pre}.

In each of the NLP tasks, the dataset used for the intermediate task training from the LLM is at least an order of magnitude larger than the dataset used for fine-tuning the model to the target task. The target task for domain adaptation, and the respective datasets, have been chosen to be in a similar field, as of the intermediate task dataset. For cross-lingual target tasks, we have tried to ensure that the task is in the same domain, and the language has semantic and syntactic similarity.

\subsection{Text classification} 
For text classification, we performed intermediate task training using the IMDB movie reviews dataset \cite{maas-EtAl:2011:ACL-HLT2011}. The IMDB movie reviews dataset has 50,000 examples classifying the movie reviews into two classes: positive and negative.

To evaluate domain adaptation, we used a randomly sampled subset of SMS spam collection dataset, with 5,600 examples \cite{delany2012sms}, annotated as spam or not. For cross-lingual task evaluation, we created a dataset of French and Spanish movie reviews, with 100 examples of each language, obtained by web scraping. Each example was translated into English using Google Translate and annotated as positive or negative.

\subsection{Sentiment analysis}
IMDB genre classification dataset was used for intermediate task training for the sentiment analysis task \cite{kumar2022movie}. This dataset consisted of 49,000 examples, classified into 27 different movie genres.

We sampled the GoEmotions dataset for 5,200 examples across all 28 different emotions, including neutral \cite{demszky2020goemotions} for the domain adaptation task. Similar to text classification, we manually created a dataset of French and German sentences with 1,700 examples, and used Google Translate to translate them into English, and then manually annotate them into 13 different emotions.

\subsection{Sentence similarity}
We used the Paraphrase Adversaries from Word Scrambling (PAWS) dataset, containing human-labeled sentence similarity for more than 49,000 pair-wise examples, for the intermediate task training \cite{paws2019naacl}.

For domain adaptation, we collected about 2,100 stock names, listed on the New York Stock Exchange and NASDAQ-traded stocks. We sampled 4,000 exampled from the PAWS-X dataset to evaluate cross-lingual sentence similarity \cite{pawsx2019emnlp}. 

\begin{figure}[t]
\centering
\includegraphics[width=240pt]{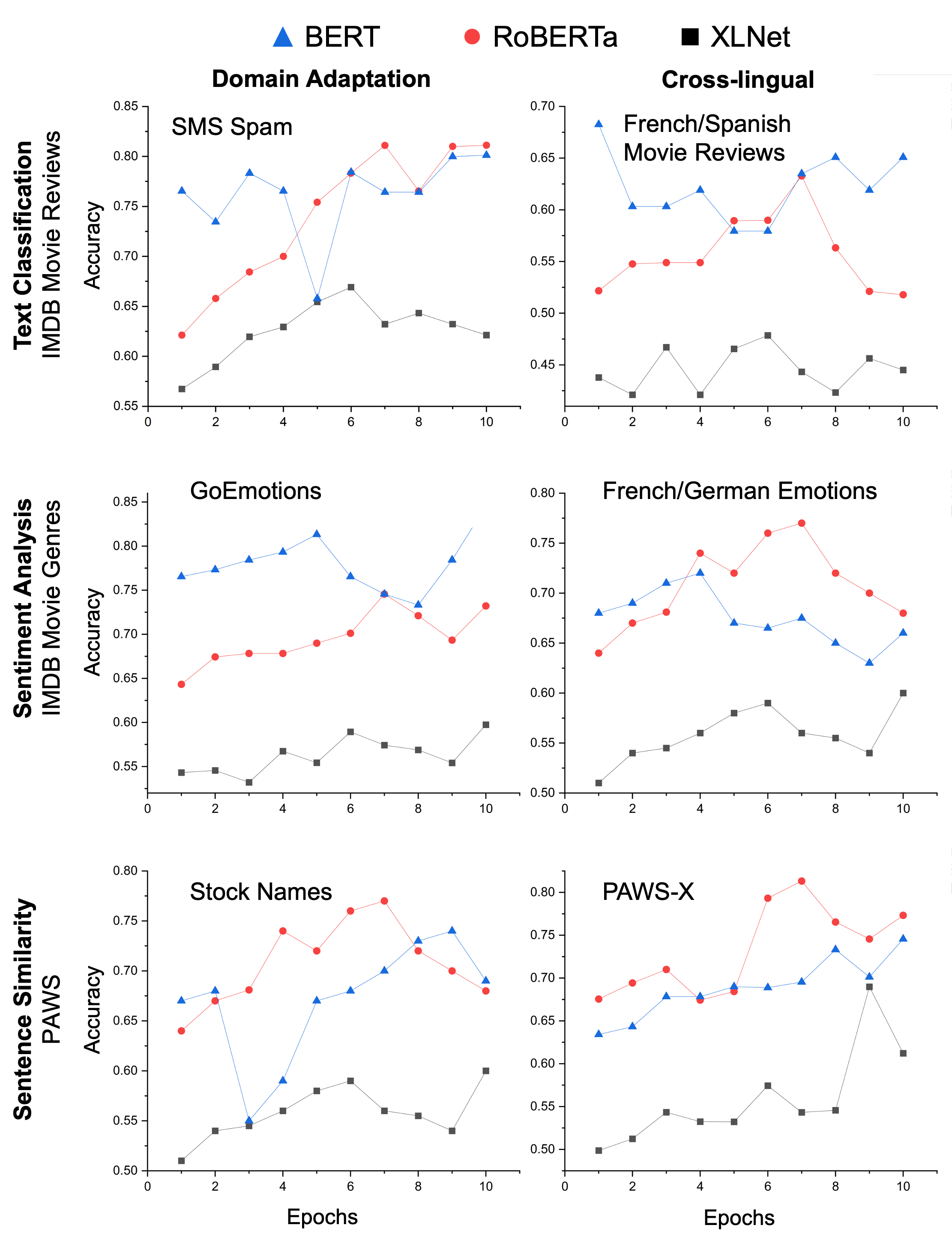}
\caption{\textbf{Accuracy on the test dataset against the fine-tuning epoch of the intermediate task transfer learnt LLMs has been noted for different tasks. }The rows represent the NLP tasks, and the columns represent the transfer learning tasks. Specific datasets used for the intermediate task training have been mentioned to the right of the NLP task, and the transfer learning datasets used for training and testing have been noted on the graphs.}
\label{fig1}
\end{figure}

\section{Results and Discussion}
To understand the effects of fine-tuning with and without intermediate task training, we analyzed the accuracy of the three NLP tasks for both domain adaptation and cross-lingual predictions. 

\subsection{RoBERTa and BERT with intermediate task training are the best models, depending on the task}
In our observations, we observed that fine-tuning a post-intermediate task training transfer learnt RoBERTa LLM outperformed in three out of six tasks, across three NLP tasks (text classification, sentiment analysis, sentence similarity) and two transfer learning experiments (domain adaptation, cross-lingual performance), while BERT outperformed others in the rest three (Figure \ref{fig1}). In both cases of text classification - domain adaptation and cross-lingual prediction, and sentiment analysis - domain adaptation, BERT outperformed RoBERTa. XLNet was consistently the worst performing model in all of our experiments. 

\begin{table*}[t]
\caption{\textbf{Accuracy of the target-task fine-tuned large language models with and without intermediate task training has been noted for the different NLP tasks. }The specific LLMs are BERT, RoBERTa, and XLNet, with the appended note mentioning if the specific model had an intermediate task training or not. Models with intermediate task training (I) on the larger dataset followed by fine-tuning (F) to the target task have been noted as `ModelName-IF', such as BERT-IF, and if the model has only been fine-tuned, then it has been noted as `ModelName-F', such as BERT-F. In target tasks per NLP task, the first task is for domain adaptation, and the next one is for cross-lingual adaptation. The best performing model accuracy for each NLP target-task has been noted in bold.
}
\label{tab1}
\resizebox{500pt}{!}{%
\begin{tabular}{cccccccc}
\hline
\rule{0pt}{2ex}    
\textbf{NLP Task} & \textbf{Target Task} & \textbf{BERT-IF} & \textbf{BERT-F} & \textbf{RoBERTa-IF} & \textbf{RoBERTa-F} & \textbf{XLNet-IF} & \textbf{XLNet-F} \\ \hline
\rule{0pt}{2ex}    
\multirow{2}{*}{Text Classification} & SMS Spam & 0.62 & 0.71 & 0.55 & \textbf{0.73} & 0.48 & 0.58 \\ \cline{2-8} \rule{0pt}{2ex}    

 & French/Spanish Movie Reviews & \textbf{0.67} & 0.63 & 0.51 & 0.58 & 0.46 & 0.44 \\ \hline
\rule{0pt}{2ex}    
\multirow{2}{*}{Sentimental Analysis} & GoEmotions & \textbf{0.89} & 0.83 & 0.76 & 0.71 & 0.62 & 0.59 \\ \cline{2-8} \rule{0pt}{2ex}    
 & French/German Emotions & 0.61 & 0.68 & 0.66 & \textbf{0.72} & 0.53 & 0.57 \\ \hline
\rule{0pt}{2ex}    
\multirow{2}{*}{Sentence Similarity} \rule{0pt}{2ex}    
& Stock Ticker & 0.52 & 0.67 & 0.67 & \textbf{0.72} & 0.47 & 0.55 \\ \cline{2-8} \rule{0pt}{2ex}  
 & PAWS-X & 0.61 & 0.67 & 0.71 & \textbf{0.74} & 0.62 & 0.58 \\ \hline
\end{tabular}%
}
\end{table*}

Similar trends for transfer learning using LLMs, where RoBERTa and BERT have similar performance, and both outperform XLNet, have been recently reported for detection of clickbaits \cite{rajapaksha2021bert}, fake reviews \cite{gupta2021leveraging}, and mental illness prediction \cite{ameer2022mental}. 

Irrespective of the tasks, we noted that the change in accuracy against the fine-tuning epochs for the LLMs the followed a similar trend. RoBERTa and XLNet improved on the accuracy metric for about 5 to 7 epochs, and then the accuracy started to go down. BERT had almost similar accuracy with minor improvements, or even worsening, as the fine-tuning epochs increased. We attribute such behavior to an observation of the linguistic information storage noted by pruning of layers in BERT, RoBERTa, and XLNet in a recent work \cite{durrani2021transfer}, stating that BERT stored the information deep in the network, while RoBERTa and XLNet localized it in the lower layers. Their observations qualitatively indicate higher rates of fine-tuning in the early epochs for RoBERTa and XLNet, aligning with our observations. In the later epochs, such behavior aligns with the bias-variance trade-off that comes with fine-tuning a generalized model to the target task, where too much training erodes the generalizability of the intermediate task training and starts overfitting to the target dataset. 

\subsection{Fine-tuned RoBERTa, without intermediate task training, outperforms other transfer learnt LLMs in most tasks}
Across all models, both with and without intermediate task training, RoBERTa, only fine-tuned on the target dataset, has the highest accuracy in four out of six tasks, while BERT with intermediate task training followed by fine-tuning leads in the other two - text classification - cross-lingual prediction and sentiment analysis - domain adaptation  (Table \ref{tab1}). Comparison of accuracy with and without intermediate task training for the LLMs provides us the opportunity to discuss the effects of positive and negative transfer, and thus the (in)effectiveness of transfer learning for domain adaptation and cross-lingual predictions.

The negative transfer noted across most of the tasks indicated that intermediate task training led to an unnecessary and ineffective generalization for the domain adaptation and cross-lingual tasks. For instance, in the case of text classification - domain adaptation, the intermediate task training on IMDB movie reviews, followed by fine-tuning, led to the best model as BERT with an accuracy of 0.62, whereas direct fine-tuning led to an accuracy of 0.71, 0.73, and 0.58 for BERT, RoBERTa, and XLNet, respectively. The similar performance of BERT and RoBERTa, in this specific task, show how fine-tuning is effective. The performance of XLNet, with an accuracy of 0.58, which is closer to the best intermediate task trained - fine-tuned model's accuracy of 0.62, further provides convincing argument for the case. We observed similar trends for the negative transfer for other tasks, except in the cases where BERT with both intermediate task training and fine-tuning on the target dataset outperformed the only fine-tuned models. We further do a case-by-case analysis and try to provide a qualitative argument for the positive and negative transfer.

For text classification, we noted that the SMS spam collection, though seemingly similar to the movie reviews, is fundamentally different with distinct spamming patterns, thereby leading to better performance on direct fine-tuning. However, cross-lingual predictions for French/Spanish movie reviews follow a linguistically, specifically the semantics and syntax, similar notation to the English reviews, thereby leading to a positive transfer with the intermediate task training.  

In the context of sentiment analysis, sentiments in GoEmotions dataset are similar to the movie genres, for instance, the emotion \textit{love} corresponding to the genre \textit{romance}, which might be leading to an internal covariate shift with the fine-tuning, preceded by the intermediate task training, thus being aided by a positive transfer. However, the French-German emotions classification have only thirteen emotions, instead of the twenty-seven genres and twenty-eight emotions in the GoEmotions domain adaptation task, which might lead to a harder covariate shift in the confusion matrix for classification. Additionally, it shows that the syntactic knowledge of the English language might not be the only factor in understanding the
emotion behind the French/German text. 

Ultimately, for sentence similarity, we attribute the directly fine-tuned model performing better on the financial domain adaptation to the lack of inherent pattern for the full names of tickers, which have been adapted from a very specialized domain. In this case, the linguistic knowledge from the general PAWS dataset does not transfer any finance-specific knowledge to the model. We believe that a similar argument can be made for the confusing nature of syntax in any particular language used when paraphrasing, thereby leading to a negative transfer for a cross-lingual sentence similarity task.

\section{Limitations and Future Work}
In the case of domain adaptation, the focus on SMS spam similarity to movie reviews, genre to emotions, is a limited field of view. We would like to expand on these tasks and extend the study to multiple cases to see if there are design principles based on similarity of the domain that could predict positive or negative transfer. We believe that future work including multiple domains, with a quantitative similarity on the domains themselves, would be helpful in arriving at conclusions about the effectiveness of transfer learning in domain adaptation.

For transfer learning for cross-lingual tasks, we are limited by the small lexicons considering closely related languages to English. Furthermore, the the model performance on the cross-lingual tasks is dependent on the task and the diversity in the source data. Considering unrelated languages, such as Chinese, Korean, Hindi and more would shed light on how transfer learning strategies could be developed to provide better predictions, without the necessity of training models from scratch. 

\section{Conclusion}
In this work, we analyzed the effectiveness of transfer learning by fine-tuning to a target task, with and without an intermediate task training step. We observed that in almost all cases, the intermediate task training leads to a negative transfer for both domain adaptation and cross-lingual NLP tasks. On the performance of different LLMs, we noted that BERT and RoBERTa performed similarly, and outperformed XLNet on the tasks when an intermediate task training step was involved. However, when we compared all models both with and without the intermediate task training step, only fine-tuned RoBERTa emerged as a clear winner. We hope that these results, and those that arise as a part of this work in the near future, inform the community on the positive and negative effects of transfer learning with an intermediate task training step.

\bibliography{aaai22}

\begin{thebibliography}{37}
\providecommand{\natexlab}[1]{#1}

\bibitem[{Ahmad et~al.(2020)Ahmad, Jindal, Ekbal, and
  Bhattachharyya}]{ahmad2020borrow}
Ahmad, Z.; Jindal, R.; Ekbal, A.; and Bhattachharyya, P. 2020.
\newblock Borrow from rich cousin: transfer learning for emotion detection
  using cross lingual embedding.
\newblock \emph{Expert Systems with Applications}, 139: 112851.

\bibitem[{Ameer et~al.(2022)Ameer, Arif, Sidorov, Gomez-Adorno, and
  Gelbukh}]{ameer2022mental}
Ameer, I.; Arif, M.; Sidorov, G.; Gomez-Adorno, H.; and Gelbukh, A. 2022.
\newblock Mental illness classification on social media texts using deep
  learning and transfer learning.
\newblock \emph{arXiv preprint arXiv:2207.01012}.

\bibitem[{Baroiu and Traușan-Matu(2022)}]{buaroiu2022automatic}
Baroiu, A.-C.; and Traușan-Matu, S. 2022.
\newblock Automatic Sarcasm Detection: Systematic Literature Review.
\newblock \emph{Information}, 13(8): 399.

\bibitem[{Bengio(2012)}]{bengio2012deep}
Bengio, Y. 2012.
\newblock Deep learning of representations for unsupervised and transfer
  learning.
\newblock In \emph{Proceedings of ICML workshop on unsupervised and transfer
  learning}, 17--36. JMLR Workshop and Conference Proceedings.

\bibitem[{Chen et~al.(2018)Chen, Awadallah, Hassan, Wang, and
  Cardie}]{chen2018multi}
Chen, X.; Awadallah, A.~H.; Hassan, H.; Wang, W.; and Cardie, C. 2018.
\newblock Multi-source cross-lingual model transfer: Learning what to share.
\newblock \emph{arXiv preprint arXiv:1810.03552}.

\bibitem[{Day and Khoshgoftaar(2017)}]{day2017survey}
Day, O.; and Khoshgoftaar, T.~M. 2017.
\newblock A survey on heterogeneous transfer learning.
\newblock \emph{Journal of Big Data}, 4(1): 1--42.

\bibitem[{Delany, Buckley, and Greene(2012)}]{delany2012sms}
Delany, S.~J.; Buckley, M.; and Greene, D. 2012.
\newblock SMS spam filtering: Methods and data.
\newblock \emph{Expert Systems with Applications}, 39(10): 9899--9908.

\bibitem[{Dell’Orletta(2009)}]{dell2009ensemble}
Dell’Orletta, F. 2009.
\newblock Ensemble system for Part-of-Speech tagging.
\newblock \emph{Proceedings of EVALITA}, 9: 1--8.

\bibitem[{Demszky et~al.(2020)Demszky, Movshovitz-Attias, Ko, Cowen, Nemade,
  and Ravi}]{demszky2020goemotions}
Demszky, D.; Movshovitz-Attias, D.; Ko, J.; Cowen, A.; Nemade, G.; and Ravi, S.
  2020.
\newblock GoEmotions: A dataset of fine-grained emotions.
\newblock \emph{arXiv preprint arXiv:2005.00547}.

\bibitem[{Durrani, Sajjad, and Dalvi(2021)}]{durrani2021transfer}
Durrani, N.; Sajjad, H.; and Dalvi, F. 2021.
\newblock How transfer learning impacts linguistic knowledge in deep NLP
  models?
\newblock \emph{arXiv preprint arXiv:2105.15179}.

\bibitem[{Felbo et~al.(2017)Felbo, Mislove, S{\o}gaard, Rahwan, and
  Lehmann}]{felbo2017using}
Felbo, B.; Mislove, A.; S{\o}gaard, A.; Rahwan, I.; and Lehmann, S. 2017.
\newblock Using millions of emoji occurrences to learn any-domain
  representations for detecting sentiment, emotion and sarcasm.
\newblock \emph{arXiv preprint arXiv:1708.00524}.

\bibitem[{Ghafoorian et~al.(2017)Ghafoorian, Mehrtash, Kapur, Karssemeijer,
  Marchiori, Pesteie, Guttmann, Leeuw, Tempany, Ginneken
  et~al.}]{ghafoorian2017transfer}
Ghafoorian, M.; Mehrtash, A.; Kapur, T.; Karssemeijer, N.; Marchiori, E.;
  Pesteie, M.; Guttmann, C.~R.; Leeuw, F.-E.~d.; Tempany, C.~M.; Ginneken,
  B.~v.; et~al. 2017.
\newblock Transfer learning for domain adaptation in MRI: Application in brain
  lesion segmentation.
\newblock In \emph{International conference on medical image computing and
  computer-assisted intervention}, 516--524. Springer.

\bibitem[{Gupta, Gandhi, and Chakravarthi(2021)}]{gupta2021leveraging}
Gupta, P.; Gandhi, S.; and Chakravarthi, B.~R. 2021.
\newblock Leveraging transfer learning techniques-bert, roberta, albert and
  distilbert for fake review detection.
\newblock In \emph{Forum for Information Retrieval Evaluation}, 75--82.

\bibitem[{Houlsby et~al.(2019)Houlsby, Giurgiu, Jastrzebski, Morrone,
  De~Laroussilhe, Gesmundo, Attariyan, and Gelly}]{houlsby2019parameter}
Houlsby, N.; Giurgiu, A.; Jastrzebski, S.; Morrone, B.; De~Laroussilhe, Q.;
  Gesmundo, A.; Attariyan, M.; and Gelly, S. 2019.
\newblock Parameter-efficient transfer learning for NLP.
\newblock In \emph{International Conference on Machine Learning}, 2790--2799.
  PMLR.

\bibitem[{Kim et~al.(2020)Kim, Kim, Cho, Song, Lee, Ahn, Park, Gong, and
  Kim}]{kim2020effectiveness}
Kim, Y.-G.; Kim, S.; Cho, C.~E.; Song, I.~H.; Lee, H.~J.; Ahn, S.; Park, S.~Y.;
  Gong, G.; and Kim, N. 2020.
\newblock Effectiveness of transfer learning for enhancing tumor classification
  with a convolutional neural network on frozen sections.
\newblock \emph{Scientific Reports}, 10(1): 1--9.

\bibitem[{Kouw and Loog(2018)}]{kouw2018introduction}
Kouw, W.~M.; and Loog, M. 2018.
\newblock An introduction to domain adaptation and transfer learning.
\newblock \emph{arXiv preprint arXiv:1812.11806}.

\bibitem[{Kumar et~al.(2022)Kumar, Kumar, Dev, and Naorem}]{kumar2022movie}
Kumar, S.; Kumar, N.; Dev, A.; and Naorem, S. 2022.
\newblock Movie genre classification using binary relevance, label powerset,
  and machine learning classifiers.
\newblock \emph{Multimedia Tools and Applications}, 1--24.

\bibitem[{Luo et~al.(2021)Luo, Wang, Cheng, Xiao, Xiao, Kucsko, O’Neill,
  Balam, Deng, Flores et~al.}]{luo2021cross}
Luo, J.; Wang, J.; Cheng, N.; Xiao, E.; Xiao, J.; Kucsko, G.; O’Neill, P.;
  Balam, J.; Deng, S.; Flores, A.; et~al. 2021.
\newblock Cross-language transfer learning and domain adaptation for end-to-end
  automatic speech recognition.
\newblock In \emph{2021 IEEE International Conference on Multimedia and Expo
  (ICME)}, 1--6. IEEE.

\bibitem[{Maas et~al.(2011)Maas, Daly, Pham, Huang, Ng, and
  Potts}]{maas-EtAl:2011:ACL-HLT2011}
Maas, A.~L.; Daly, R.~E.; Pham, P.~T.; Huang, D.; Ng, A.~Y.; and Potts, C.
  2011.
\newblock Learning Word Vectors for Sentiment Analysis.
\newblock In \emph{Proceedings of the 49th Annual Meeting of the Association
  for Computational Linguistics: Human Language Technologies}, 142--150.
  Portland, Oregon, USA: Association for Computational Linguistics.

\bibitem[{Meftah et~al.(2021)Meftah, Semmar, Tamaazousti, Essafi, and
  Sadat}]{meftah2021hidden}
Meftah, S.; Semmar, N.; Tamaazousti, Y.; Essafi, H.; and Sadat, F. 2021.
\newblock On the hidden negative transfer in sequential transfer learning for
  domain adaptation from news to tweets.
\newblock In \emph{Proceedings of the Second Workshop on Domain Adaptation for
  NLP}, 140--145.

\bibitem[{Mozafari, Farahbakhsh, and Crespi(2019)}]{mozafari2019bert}
Mozafari, M.; Farahbakhsh, R.; and Crespi, N. 2019.
\newblock A BERT-based transfer learning approach for hate speech detection in
  online social media.
\newblock In \emph{International Conference on Complex Networks and Their
  Applications}, 928--940. Springer.

\bibitem[{Pan(2020)}]{pan2020transfer}
Pan, S.~J. 2020.
\newblock Transfer learning.
\newblock \emph{Learning}, 21: 1--2.

\bibitem[{Peng, Yan, and Lu(2019)}]{peng2019transfer}
Peng, Y.; Yan, S.; and Lu, Z. 2019.
\newblock Transfer learning in biomedical natural language processing: an
  evaluation of BERT and ELMo on ten benchmarking datasets.
\newblock \emph{arXiv preprint arXiv:1906.05474}.

\bibitem[{Pruksachatkun et~al.(2020)Pruksachatkun, Phang, Liu, Htut, Zhang,
  Pang, Vania, Kann, and Bowman}]{pruksachatkun2020intermediate}
Pruksachatkun, Y.; Phang, J.; Liu, H.; Htut, P.~M.; Zhang, X.; Pang, R.~Y.;
  Vania, C.; Kann, K.; and Bowman, S.~R. 2020.
\newblock Intermediate-task transfer learning with pretrained models for
  natural language understanding: When and why does it work?
\newblock \emph{arXiv preprint arXiv:2005.00628}.

\bibitem[{Qiu et~al.(2020)Qiu, Sun, Xu, Shao, Dai, and Huang}]{qiu2020pre}
Qiu, X.; Sun, T.; Xu, Y.; Shao, Y.; Dai, N.; and Huang, X. 2020.
\newblock Pre-trained models for natural language processing: A survey.
\newblock \emph{Science China Technological Sciences}, 63(10): 1872--1897.

\bibitem[{Raffel et~al.(2020)Raffel, Shazeer, Roberts, Lee, Narang, Matena,
  Zhou, Li, Liu et~al.}]{raffel2020exploring}
Raffel, C.; Shazeer, N.; Roberts, A.; Lee, K.; Narang, S.; Matena, M.; Zhou,
  Y.; Li, W.; Liu, P.~J.; et~al. 2020.
\newblock Exploring the limits of transfer learning with a unified text-to-text
  transformer.
\newblock \emph{J. Mach. Learn. Res.}, 21(140): 1--67.

\bibitem[{Rajapaksha, Farahbakhsh, and Crespi(2021)}]{rajapaksha2021bert}
Rajapaksha, P.; Farahbakhsh, R.; and Crespi, N. 2021.
\newblock BERT, XLNet or RoBERTa: The Best Transfer Learning Model to Detect
  Clickbaits.
\newblock \emph{IEEE Access}, 9: 154704--154716.

\bibitem[{Rayson et~al.(2007)Rayson, Archer, Baron, Culpeper, and
  Smith}]{rayson2007tagging}
Rayson, P.; Archer, D.~E.; Baron, A.; Culpeper, J.; and Smith, N. 2007.
\newblock Tagging the Bard: Evaluating the accuracy of a modern POS tagger on
  Early Modern English corpora.
\newblock In \emph{Proceedings of the Corpus Linguistics conference: CL2007}.

\bibitem[{Salza et~al.(2022)Salza, Schwizer, Gu, and
  Gall}]{salza2022effectiveness}
Salza, P.; Schwizer, C.; Gu, J.; and Gall, H.~C. 2022.
\newblock On the effectiveness of transfer learning for code search.
\newblock \emph{IEEE Transactions on Software Engineering}.

\bibitem[{Savini and Caragea(2022)}]{savini2022intermediate}
Savini, E.; and Caragea, C. 2022.
\newblock Intermediate-task transfer learning with BERT for sarcasm detection.
\newblock \emph{Mathematics}, 10(5): 844.

\bibitem[{Vlad et~al.(2019)Vlad, Tanase, Onose, and Cercel}]{vlad2019sentence}
Vlad, G.-A.; Tanase, M.-A.; Onose, C.; and Cercel, D.-C. 2019.
\newblock Sentence-level propaganda detection in news articles with transfer
  learning and BERT-BiLSTM-capsule model.
\newblock In \emph{Proceedings of the second workshop on natural language
  processing for internet freedom: Censorship, Disinformation, and Propaganda},
  148--154.

\bibitem[{Wang et~al.(2019)Wang, Li, Ye, Long, and Wang}]{wang2019transferable}
Wang, X.; Li, L.; Ye, W.; Long, M.; and Wang, J. 2019.
\newblock Transferable attention for domain adaptation.
\newblock In \emph{Proceedings of the AAAI Conference on Artificial
  Intelligence}, volume~33, 5345--5352.

\bibitem[{Xu, He, and Shu(2020)}]{xu2020transfer}
Xu, W.; He, J.; and Shu, Y. 2020.
\newblock Transfer learning and deep domain adaptation.
\newblock \emph{Advances and Applications in Deep Learning}, 45.

\bibitem[{Yang et~al.(2019)Yang, Zhang, Tar, and Baldridge}]{pawsx2019emnlp}
Yang, Y.; Zhang, Y.; Tar, C.; and Baldridge, J. 2019.
\newblock {PAWS-X: A Cross-lingual Adversarial Dataset for Paraphrase
  Identification}.
\newblock In \emph{Proc. of EMNLP}.

\bibitem[{You et~al.(2020)You, Kou, Long, and Wang}]{you2020co}
You, K.; Kou, Z.; Long, M.; and Wang, J. 2020.
\newblock Co-tuning for transfer learning.
\newblock \emph{Advances in Neural Information Processing Systems}, 33:
  17236--17246.

\bibitem[{Zhang et~al.(2022)Zhang, Deng, Zhang, and Wu}]{zhang2022survey}
Zhang, W.; Deng, L.; Zhang, L.; and Wu, D. 2022.
\newblock A survey on negative transfer.
\newblock \emph{IEEE/CAA Journal of Automatica Sinica}, 9: 1--25.

\bibitem[{Zhang, Baldridge, and He(2019)}]{paws2019naacl}
Zhang, Y.; Baldridge, J.; and He, L. 2019.
\newblock {PAWS: Paraphrase Adversaries from Word Scrambling}.
\newblock In \emph{Proc. of NAACL}.

\end{thebibliography}

\end{document}